\documentclass[11pt]{article}
\usepackage{acl2023}
\usepackage{times}
\usepackage{latexsym}
\usepackage[T1]{fontenc}
\usepackage[utf8]{inputenc}
\usepackage{microtype}
\usepackage{inconsolata}
\usepackage{graphicx}
\usepackage{amsmath}
\usepackage{booktabs}
\usepackage{url}
\usepackage{svg}
\usepackage{enumitem}
\usepackage{amssymb}
\usepackage{makecell}
\usepackage{tabularray}
\usepackage{xcolor}
\usepackage{colortbl}

\title{Less is More: The Effectiveness of Compact Typological Language Representations}

\author{
 \textbf{York Hay Ng\textsuperscript{†}},
 \textbf{Phuong Hanh Hoang\textsuperscript{†}},
 \textbf{En-Shiun Annie Lee\textsuperscript{†,‡}}
\\
 \textsuperscript{†}University of Toronto,
 \textsuperscript{‡}Ontario Tech University
 \\
 \texttt{\{york.ng,fiona.hoang\}@mail.utoronto.ca}
}
\date{}

\begin{document}
\maketitle


\begin{abstract}
Linguistic feature datasets such as URIEL+ are valuable for modelling cross-lingual relationships, but their high dimensionality and sparsity, especially for low-resource languages, limit the effectiveness of distance metrics. We propose a pipeline to optimize the URIEL+ typological feature space by combining feature selection and imputation, producing compact yet interpretable typological representations. We evaluate these feature subsets on linguistic distance alignment and downstream tasks, demonstrating that reduced-size representations of language typology can yield more informative distance metrics and improve performance in multilingual NLP applications.

\end{abstract}

\section{Introduction}
The success of cross-lingual transfer in NLP often hinges on understanding relationships between languages \cite{lin-etal-2019-langrank}. Resources such as URIEL \cite{littell-etal-2017-uriel} provide a large repository of linguistic features (typological, geographical, phylogenetic) for thousands of languages, encapsulating language properties in vector form. URIEL has been widely used to supply language features and vector distances to multilingual models \citep{lin-etal-2019-langrank, adilazuarda-etal-2024-lingualchemy, anugraha2024proxylm}. \citet{khan-etal-2025-uriel-plus} introduced \textbf{URIEL+}, extending URIEL’s typological coverage to 4555 languages and addressing usability issues.

While URIEL+ represents a significant expansion in scope, its utility is constrained by the nature of its feature space. The typological set has grown to 800 features, many of which overlap or correlate strongly, as multiple sources contribute similar information \citep{littell-etal-2017-uriel, khan-etal-2025-uriel-plus}. Furthermore, the data matrix remains sparse: after integrating five databases, 87\% of values are still missing. Previous approaches have typically treated these features as equally informative, implicitly assuming uniform weight when computing language distances. This overlooks the high dimensionality and redundancy, which can introduce noise, risk model overfitting, and skew similarity metrics with uninformative signals.


Feature selection addresses these challenges by identifying and prioritizing the most statistically and linguistically informative features. This process mitigates the “curse of dimensionality” and can improve the separability of languages \cite{kohavi1997wrappers, BELLOTTI2014115}. In high-dimensional settings, correlation-based methods have been able to achieve a significant reduction in dimensionality, eliminating more than half of all characteristics, with minimal loss of predictive performance \citep{hall1999correlation}. Such focused selection can also act as a fast, scalable pre-processing step, substantially accelerating downstream tasks \cite{ferreira2012efficient}. Applied in conjunction with missing-value imputation, this approach enhances data quality and generalization by focusing models on a core set of salient features \cite{liu2020feature}.



\begin{figure*}[t]
\centering
\includegraphics[width=\textwidth]{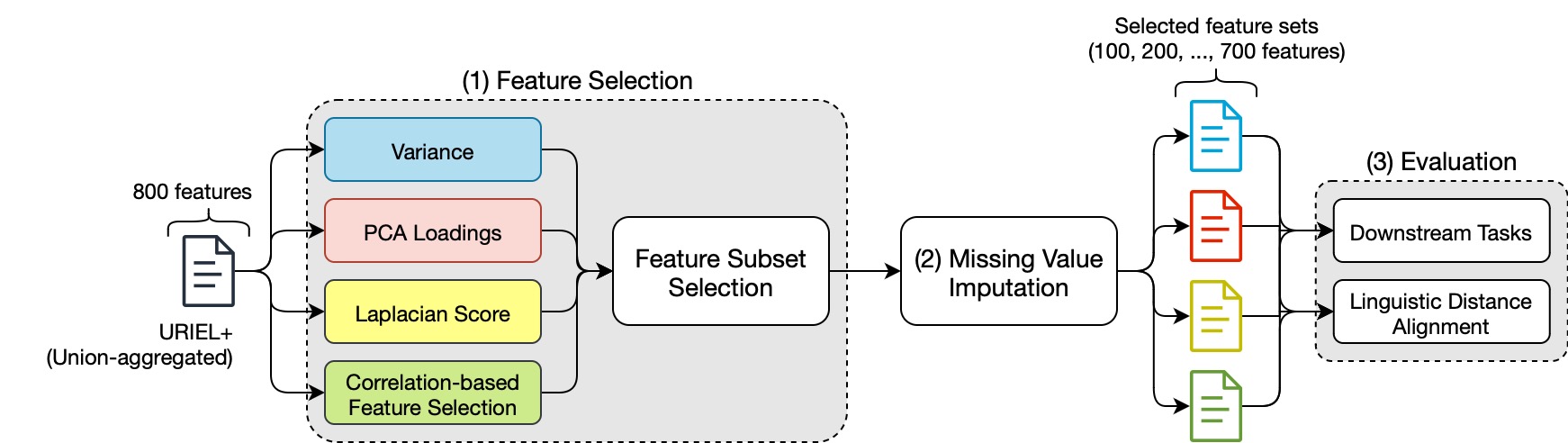}
\caption{Overview of our optimization pipeline.}
\end{figure*}

We investigate optimizing the URIEL+ typological feature set with a pipeline of \textit{feature selection} and \textit{imputation} to improve its effectiveness for multilingual NLP, filling a key gap between the growing size of URIEL+ and the need for interpretable, task-driven feature sets. We move beyond the assumption of uniform feature importance and ask: \textit{how can we produce compact, yet interpretable, typological language representations, and does this principled dimensionality reduction yield more meaningful language distances?}

Our contributions are: (1) the first principled framework for analyzing and selecting typological features, moving beyond treating all features as equally informative; (2) a robust imputation strategy with SoftImpute \cite{mazumder-etal-2010-softimpute} to handle missing values within these compact representations; and (3) an empirical demonstration that these feature subsets improve linguistic distance alignment and downstream task performance.

\section{Methodology}
Our optimization pipeline consists of three stages: (1) feature selection by four methods, including a novel approach of integrating language genetic data as class labels, (2) imputation of missing values, and (3) evaluation of the performance impact of each new feature subset. We perform our experiments on the URIEL+ \emph{typological} dataset \citep{khan-etal-2025-uriel-plus}, a three-dimensional array of \emph{languages} (4555), \emph{features} (800), and \emph{data sources} (14). We pre-process the dataset by taking union aggregation over sources (meaning a value is taken from any available source), producing a dataset of size $(4555, 800)$ containing \textit{feature column vectors} $[\mathbf{f_0},...,\mathbf{f_{799}}]$ where $\mathbf{f_i} \in \{0,1\}^{4555}$.


\subsection{Unsupervised Feature Selection}
We first employ three unsupervised filter-based methods to learn feature subsets. We evaluate each feature individually based on the statistical properties of the data, without the presence of class labels \cite{BELLOTTI2014115}. As we vary the subset size $k$ across $\{100, 200, ..., 700\}$, we rank features by importance scores $s_i$, $i \in [0,800)$, and pick the top-$k$ features based on the following metrics:

\textbf{Variance}: 
    Ranking features by their variance has been found to be effective in sparse datasets \cite{1598807, ferreira2012efficient}. Variance serves as a proxy for informativeness: a feature with minimal variance is either predominantly present or absent across languages, thereby offering utility in distinguishing between languages. The variance of a \textit{binary} feature $i$ is defined as:
    \begin{equation}
    s_i^{Var} = \text{Var}(f_i) = p_i (1 - p_i)
    \label{eq:variance}
    \end{equation}
    where $f_i$ is a random variable representing feature $i$ and $p_i$ is the proportion of languages possessing feature $i$.
    
    \textbf{Principal Component Analysis (PCA) Loadings}: 
    Principal Component Analysis (PCA) identifies directions of maximal variance in the data via orthogonal principal components (PCs) \cite{pca}. Each PC is a weighted combination of the original features, where the weight assigned to a feature is called its \textit{loading}. Intuitively, features with high absolute PCA loadings are those that explain the underlying structure of the typological dataset. We therefore select features that contribute most strongly to these components to retain interpretability while reducing dimensionality \cite{GUO2002123}, scoring feature $i$ by its maximum absolute loading among all PCs:
    \begin{equation}
    s_i^{PCA} = \max_{0\leq j < n} | b_{j,i} |
    \label{eq:pca}
    \end{equation}
    where $n$ is the minimum number of PCs needed to explain 95\% of the dataset variance, and $b_{j,i}$ is the loading for the $i$-th feature in the $j$-th PC. 

    \textbf{Laplacian Score}:
Laplacian Score is a graph-based criterion that ranks features by how well they preserve locality in the data manifold \cite{he2005laplacian}. The intuition is that meaningful features should vary smoothly across similar languages.

We construct a 5-nearest neighbour graph over the languages, using a heat kernel to assign edge weights based on feature-space distance. Given a graph \( G \) with weight matrix \( S \), degree matrix \( D \), and unnormalized Laplacian \( L = D - S \), we evaluate each feature \( \mathbf{f}_i \) by first centering it with respect to \( D \), producing $\tilde{\mathbf{f}}_i$, and computing its score:
\[
s_i^{LS} = \frac{\tilde{\mathbf{f}}_i^\top L \tilde{\mathbf{f}}_i}{\tilde{\mathbf{f}}_i^\top D \tilde{\mathbf{f}}_i}
\]
This score measures how much the feature varies locally relative to its overall variance. Lower-scoring features are more stable across similar languages and thus more likely to encode intrinsic typological patterns. We rank features in ascending order.

\subsection{Supervised Feature Selection}
While URIEL+'s language representations are inherently label-less, we further propose a strategy of applying language family membership as class labels. Using URIEL+ phylogenetic vectors, we encode the \textit{top-level language family} as a categorical class variable $c$, allowing us to directly investigate feature relevancy with respect to the class.

 \textbf{Correlation-based Feature Selection}: 
    We utilize a method inspired by Correlation-based Feature Selection \cite{hall1999correlation}, producing a feature subset which balances between feature-class relevance and inter-feature redundancy. A "hill-climbing" algorithm \cite{kohavi1997wrappers} is used to build our feature set: starting with an empty set, we iteratively score all remaining features and add the feature yielding the highest score to the set.
    
    We measure a feature's relevance by how well it distinguishes between top-level family membership. Specifically, we use Mutual Information (MI), a quantity used in linguistics and feature selection to measure how informative a feature is about the class \cite{bickel2010capturing, vergara2014review}. It can be expressed as:
    
\begin{equation}
I(f_i;c) \;=\; H(f_i) + H(c) - H(f_i,c)\,,
\label{eq:mi-entropy}
\end{equation}
where $f_i$ and $c$ are random variables representing the feature and the class, respectively, and $H(\cdot)$ denotes Shannon entropy \cite{cover2006elements}.

To mitigate redundancy between features, we scale the relevance score by the sum of correlations between the candidate feature and all features in the current subset. We measure the Phi correlation $\phi$ between features, which is ideal for binary features. The resulting merit score assigned to feature $i$ is defined as:

\begin{equation}
s_i^{CFS} = \frac{I(f_i;c)}{\sum_{\substack{j=1, j \neq i}}^n \phi(\mathbf{f_i}; \mathbf{f_j})}
\end{equation}

where $n$ is the current subset size. In each iteration of the search algorithm, this score is recomputed and determines the best feature to select.

\subsection{Missing Value Imputation}
To address missing values in the feature subsets, we apply SoftImpute \citep{mazumder-etal-2010-softimpute} to complete the language-feature matrix, as it was demonstrated to be the strongest imputation method in URIEL+ \cite{khan-etal-2025-uriel-plus}. SoftImpute fills unknown typological values in the reduced feature space by approximating a low-rank matrix, distributing information from observed entries to infer missing ones. This low-rank assumption helps preserve global structure while reducing noise. 

\begin{figure*}
    \centering
    \includegraphics[width=1\textwidth]{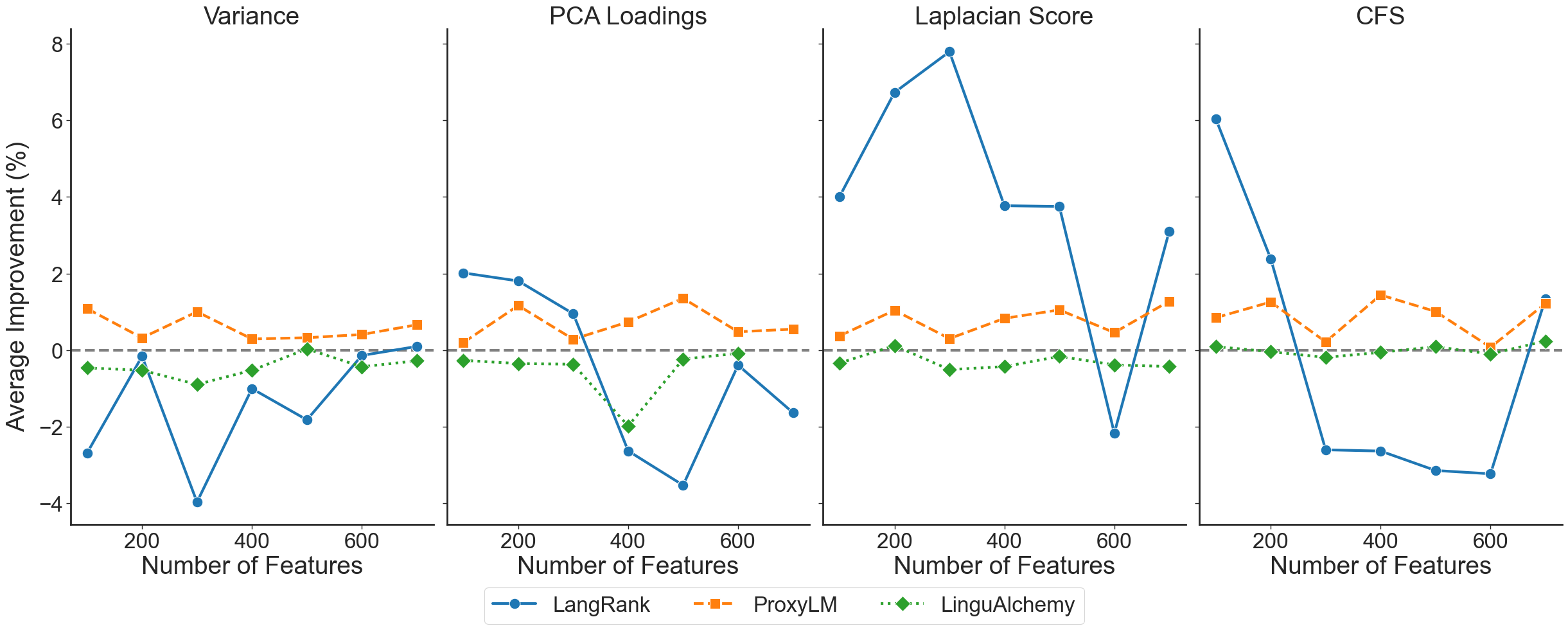}
    \caption{Average improvement (\%) in downstream task performance: \textsc{LangRank} NDCG@3, \textsc{ProxyLM} RMSE and \textsc{LinguAlchemy} accuracy, compared to applying baseline URIEL+ distances.}
    \label{fig:downstream-results}
\end{figure*}

\section{Linguistic Distance Alignment}
To assess how well typological distances derived from feature subsets reflect linguistic similarity, we evaluate the alignment between our derived distances with a known linguistics similarity measure. We consider a set of similarity scores between 28 language pairs, comprising 24 low-resource languages in the Isthmo-Colombian area, identified in a case study by \citet{hammarstrom-oconnor-2013-dependency}. These scores, derived from a modified Gower coefficient $G_d$, measure typological distance while accounting for feature dependencies.

We compute angular distances between language vectors from each feature subset, and report Spearman correlation coefficients $\rho$ between our derived distances and the known similarity score $G_d$ \footnote{Full results can be found in Appendix \ref{appendix:align}.}

\subsection{Alignment Results}
\begin{table}[h]
\centering\small
\begin{tabular}{lcc}
\toprule
Feature Set & Best Subset Size & $\rho$ w.r.t $G_d$ \\
\midrule
URIEL+ (Baseline)      & -- & 0.260 \\
\midrule
Variance             & 500 & 0.295 \\
PCA Loadings             & 700 & 0.292 \\
Laplacian Score             & 300 & 0.358 \\
Correlation FS            & 200 & 0.264 \\
\bottomrule
\end{tabular}
\caption{Exemplar alignment results for each method, in terms of Spearman correlations between language distances from feature subsets and the linguistic similarity measure $G_d$.}
\label{tab:results}
\end{table}

All feature selection methods, at the right subset size, outperforms the full URIEL+ feature space in distance alignment with linguistic reality. Although no feature subset produced distances which significantly correlated with $G_d$ at the $p=0.05$ significance level, considering the small sample size, the stronger alignments demonstrate the improved accuracy of distance-based comparisons between languages despite utilizing smaller feature sets, and suggests that our method enhances typological differentiability between languages.

Notably, the Laplacian Score strategy achieves the strongest $G_d$ alignment, with $\rho=0.358$ \\ ($p=0.061$) when using only 300 features, underscoring its effectiveness at capturing manifold structure. Simultaneously, in lower-dimensional spaces, the CFS strategy maintains similar alignment compared to URIEL+, highlighting its strength in selecting the most mutually informative features, while other methods excel in higher-dimensional spaces.

Despite the challenge of data sparsity, the consistent performance gains further indicate that our combined pipeline of feature selection and imputation can effectively trim noisy and redundant features, producing compact representations that generalize effectively for low-resource languages.

\section{Downstream Task Performance}
We test the utility of feature subsets on various NLP tasks which rely on typological features, focusing on application in multilingual tasks and models. \footnote{Detailed results and settings are shown in Appendix \ref{appendix:downstream}.}

\subsection{Cross-Lingual Transfer Prediction}
\textsc{LangRank} \cite{lin-etal-2019-langrank} is a framework for choosing transfer languages for NLP tasks requiring cross-lingual transfer based on language distances. For each feature subset we recompute angular distances between languages, grouped by type (syntax, morphology, etc.). These updated distances are supplied to \textsc{LangRank} to re-rank transfer languages across four subtasks: dependency parsing (DEP), machine translation (MT), entity linking (EL), and part-of-speech tagging (POS). We measure \textsc{LangRank}'s ranking performance with average top-3 Normalized Discounted Cumulative Gain (NDCG@3).

\subsection{Language Model Performance Prediction}
\textsc{ProxyLM} \cite{anugraha2024proxylm} is a framework for predicting task performance of multilingual language models using typological distances. We replace the original distances with angular distances from each subset and train XGBoost regressors to predict performance of M2M100 \cite{fan2020englishcentricmultilingualmachinetranslation} and NLLB \cite{nllbteam2022languageleftbehindscaling} on machine translation tasks MT560 \cite{gowda-etal-2021-many} and NusaTranslation \cite{purwarianti2023nusatranslation}, evaluating with 5-fold root mean squared error (RMSE).

\subsection{Language Model Linguistic Regularization}
LinguAlchemy \cite{adilazuarda-etal-2024-lingualchemy} introduces a linguistic regularization term that aligns text representations in language models with language vectors during training. While the original study used syntax vectors (i.e. comprising only syntactic features from the typological vector), we extract syntax vectors from each feature subset to replace the original vector. We subsequently measure the accuracy of mBERT \cite{devlin2019bert} and XLM-R \cite{conneau2019unsupervised} on MasakhaNews topic classification \cite{adelani-etal-2023-masakhanews} and MASSIVE intent classification \cite{fitzgerald2022massive} after aligning text representations.

\subsection{Experimental Results}

Figure~\ref{fig:downstream-results} shows the average improvements in task performance across feature subset sizes (100–700) for each feature selection method, compared to URIEL+ distances and vectors.

For \textsc{LangRank}, the LS and CFS methods outperform the full feature set baseline at smaller subset sizes, with peak gains of +7.8\% (size 300) and +6.0\% (size 100) respectively. This highlights the utility of languages distances from these methods in choosing better transfer languages, while demonstrating how they excel in smaller subsets for capturing defining typological features. By contrast, Variance and PCA Loading strategies offer less improvement, indicating that feature variance and PC contribution alone are less effective at preserving structural information.

For \textsc{ProxyLM}, performance gains are smaller but consistent across all subset sizes and selection methods, with relative RMSE decrease of 0.2\% to 1.4\%. Crucially, no subset results in performance degradation. This suggests that, across subset sizes, our pipeline produces robust distances which contain sufficient information for predicting language model performance.

In comparison, when aligning text representations in language models to typological language vectors under the \textsc{LinguAlchemy} framework, performance varies only slightly, suggesting that the regularization loss is robust to feature reduction: as long as broad syntactic cues are preserved, smaller vectors provide signals comparable to the full set. This further supports the idea that compact representations suffice, with little benefit from carrying redundant features.

Overall, these results support our central hypothesis: principled feature selection and imputation reduces redundancy in typological datasets, improving the utility of distance values. Although different feature selection methods and subset sizes impact different downstream tasks uniquely, the fact that low-dimensional feature spaces from all four methods performs at least comparably to the 800-dimensional baseline in aiding \textsc{LangRank} and \textsc{ProxyLM} prediction suggests that reduced-dimensionality feature subsets better represent typological distance between languages, yielding improved predictive performance in downstream tasks while enabling more runtime-efficient distance computations.

\section{Conclusion}
We present a pipeline for producing compact, interpretable representations of typological features from the URIEL+ database. Through four feature selection strategies and imputation, we derived feature sets that improve the informativeness of typological distances even for low-resource languages, despite the dimensionality reduction of the feature space. Empirical results demonstrate their enhanced utility: new vectors improve typological comparability, alongside cross-lingual transfer and language model performance prediction accuracy, while reducing runtime costs. Our findings therefore suggest that \textit{“less is more”} when modelling language relationships with typological features in multilingual NLP.


\section*{Limitations}
\textbf{Scalar distance.} While our evaluation of feature subsets utilizes angular distances to quantify language similarity, any attempt to collapse complex linguistic relationships (encompassing syntax, phonology, and typology) into a single scalar distance inevitably oversimplifies. Different linguistic dimensions may require distinct representational strategies, and a unified distance metric cannot fully capture this richness.

\textbf{Biases in feature selection.} Each feature selection method inherently favors different properties: for instance, PCA favors features contributing to global variance, while CFS emphasizes local mutual relevance. As such, selected subsets may emphasize certain linguistic properties over others. A hybrid or ensemble-based selection strategy may yield more balanced representations.

\textbf{Imputation techniques.} We rely solely on SoftImpute for missing-value imputation. While it performs well under low-rank assumptions, alternative techniques (e.g., random forest models or autoencoders) could better capture linguistic relationships and offer complementary benefits, particularly for highly sparse languages or feature types.

\textbf{Language coverage.} In comparison to the 4,555 languages covered by URIEL+'s typological dataset, our evaluation focuses on a relatively small set of languages (19 in the alignment study, 105 in \textsc{LangRank}, 51 in \textsc{ProxyLM} and 63 in \textsc{LinguAlchemy}), with an emphasis on high-resource languages. While informative, this limits the generalizability of our findings across language families and resource levels. As seen in our results, our method achieves varying success across downstream applications. Future work should incorporate broader benchmarks, including diverse downstream tasks and typologically varied evaluation sets.

\section*{Ethics Statement}
This work uses typological features which are derived from publicly available linguistic databases, and does not involve personal or sensitive data. Representing languages as feature vectors necessarily simplifies complex linguistic realities and may reflect biases in the source data, which users should keep in mind. While our contribution is primarily methodological and removed from direct applications, cross-lingual NLP carries both benefits (e.g., supporting low-resource languages) and risks of misuse. To support transparency and responsible use, we release our code publicly (Appendix \ref{appendix:reproduce}).

\section*{Acknowledgements}
We would like to thank Mason Shipton and Jun Bin Cheng for their contributions, along with Giancarlo Giannetti, Aron-Seth Cohen and Bridget Green for the initiation of this study.

\bibliographystyle{acl_natbib}
\bibliography{main}

\appendix

\section{Related Work}
\paragraph{Linguistic Feature Bases: URIEL and URIEL+.}
\citet{littell-etal-2017-uriel} created the URIEL knowledge base to represent languages as vectors of linguistic features. These include typological features (drawn largely from the World Atlas of Language Structures and similar sources), phylogenetic features (language family indicators), and geographical coordinates, which are drawn from Glottolog \cite{glottolog}. URIEL also defined distance metrics over these vectors, enabling calculation of inter-language distances \citep{littell-etal-2017-uriel}.

URIEL+ \citep{khan-etal-2025-uriel-plus} is an enhanced version that expands feature coverage and addresses limitations. It integrates additional databases (e.g. SAPhon for phonology, Grambank for morphology) to increase the number of features and languages covered. URIEL+ also introduces multiple imputation methods for missing data (including $k$-NN, MIDASpy, and SoftImpute) and allows customizable distance calculations. These changes improved downstream task performance and produced language distance estimates more aligned with linguistic reality \citep{khan-etal-2025-uriel-plus}.

Earlier work has sought to mitigate sparsity and redundancy in typological vectors. \citet{murawaki-2015-continuous} used autoencoders to project WALS features into continuous latent spaces for phylogenetic inference, while \citet{oncevay-etal-2020-bridging} fused typological vectors with task-learned embeddings to improve multilingual MT transfer.

However, prior to our work, the issue of redundancy and high dimensionality in typological vectors had not been explicitly addressed without sacrificing linguistic interpretability. In practice, researchers sometimes select subsets of features manually for specific tasks \citep{papadimitriou-jurafsky-2020, zhang-toral-2019, berzak-etal-2017-predicting}, indicating that a one-size-fits-all feature set may be suboptimal.

\paragraph{Feature Selection and Reduction in NLP.}
Feature selection is a well-studied problem in machine learning, including NLP tasks where high-dimensional representations are common. Traditional approaches include filtering by statistical tests (e.g. chi-square for classification relevance) or greedy elimination of features with high pairwise correlation. In multilingual settings, \citet{lin-etal-2019-langrank} learned to rank language features by importance for transfer learning. Other works have incorporated typological features into models via learned embeddings or regularization \citep{adilazuarda-etal-2024-lingualchemy, bjerva-augenstein-2018-phonology}, implicitly performing a form of dimensionality reduction by focusing on the most informative typological aspects. Our approach explicitly reduces dimensions through trimming redundant features, allowing us to retain interpretability of features.
\begin{figure*}[h]
\centering
\includegraphics[width=1\textwidth]{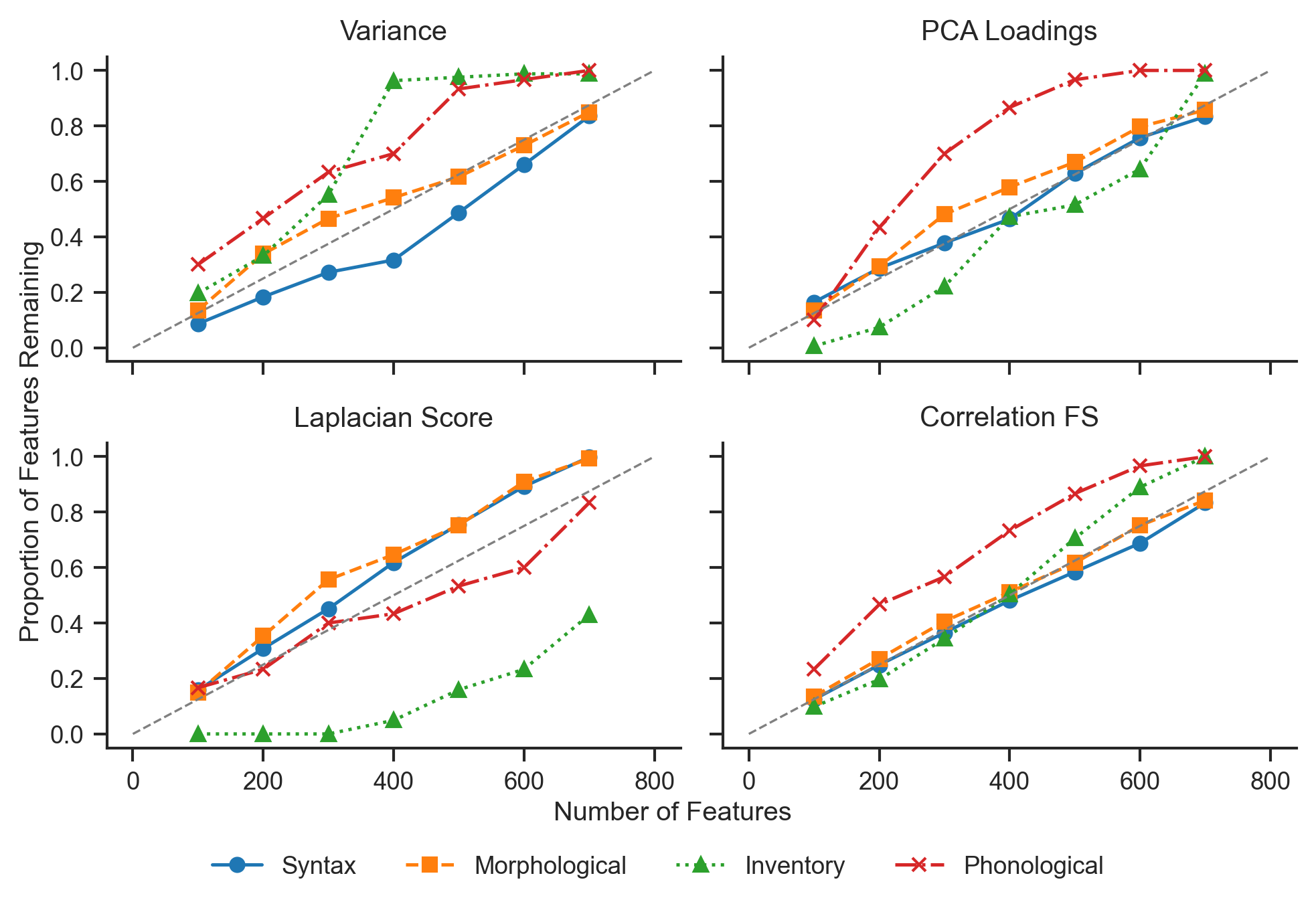}
\caption{Proportion of features remaining in all four feature subsets, by feature type. Dotted line represents expected trend if selection is type-agnostic.}
\label{fig:type-analysis}
\end{figure*}
\paragraph{Missing Data Imputation.}
Missing feature values are prevalent in typological databases. In NLP, imputation has been explored to address the challenge of sparsity in linguistic databases such as WALS. \citet{bjerva2019probabilistic} proposed using probabilistic models and language embeddings to predict missing typological values, while \citet{malaviya-etal-2017-learning} used a neural machine translation model to predict missing values in the URIEL database. SoftImpute, proposed by \citet{mazumder-etal-2010-softimpute}, is a matrix completion method that has shown the strongest performance in URIEL+ evaluations \cite{khan-etal-2025-uriel-plus}. We leverage SoftImpute to fill large portions of the URIEL+ matrix.

\section{Feature Selection Analysis}

\begin{figure}[t]
\centering
\includegraphics[width=1\columnwidth]{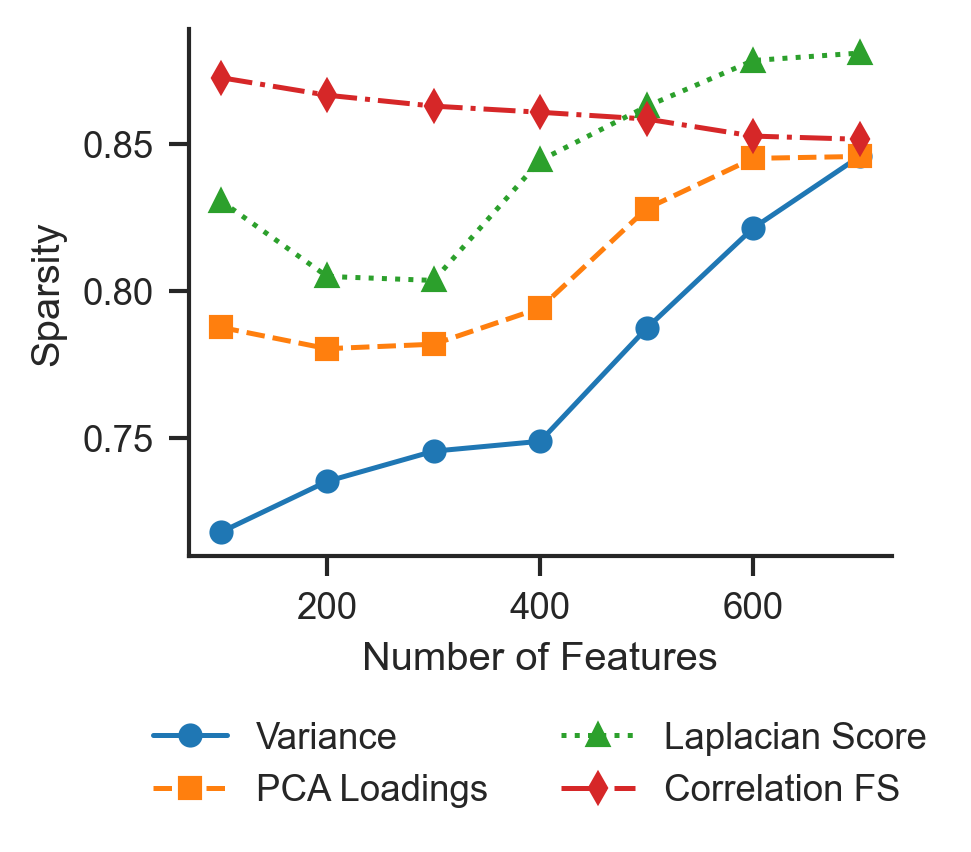}
\caption{Average sparsity (proportion of missing values) of features across all four feature subsets.}
\label{fig:sparsity-analysis}
\end{figure}

Through analyzing each feature subset, we can further investigate the linguistic or statistical properties favoured by different feature selection method.

\textbf{Feature type.} Analysis of selected feature types in each of the four subsets (Figure~\ref{fig:type-analysis}) further reveals structure in the optimized subsets. Inventory features, describing presence or absence of specific phonemes, were often excluded, especially by Laplacian and PCA-based methods, suggesting that they contribute less to preserving linguistic topology or transfer-relevant structure. On the other hand, phonological features were more frequently retained, particularly under PCA and CFS methods, implying phonological characteristics carry more discriminative power across languages in the context of multilingual NLP.

\textbf{Sparsity.} While feature selection appears to be largely agnostic to sparsity, variance-based selection tends to favour less sparse features in smaller subsets, indicating that typological features with broader coverage better balance presence and absence across languages. Correlation-based feature selection, conversely, favours \textit{sparser} features, suggesting that these features are more relevant to language phylogeny.

\section{Linguistic Alignment \label{appendix:align}}

\begin{table}[t]
\centering
\small
\begin{tblr}{
  width = \columnwidth,
  colspec = {c|cccc},
  row{1} = {font=\bfseries},
  row{2} = {font=\bfseries},
  cell{1}{1} = {r=2}{},
  cell{1}{2} = {c=4}{},
  hline{1} = {-}{1.0pt},
  hline{3} = {-}{0.8pt},
  vline{2} = {2-8}{0.5pt},
  hline{10} = {-}{0.8pt},
  hline{11} = {-}{1.0pt},
}
\SetCell[r=2]{c,m} {\begin{tabular}{@{}c@{}}\textbf{Subset}\\\textbf{Size}\end{tabular}} & \SetCell[c=4]{c} \textbf{Feature Selection Method} &  &  & \\
& \textbf{Var} & \textbf{PCA} & \textbf{LS} & \textbf{CFS} \\
100 & \textbf{0.289} & 0.245 & 0.108 & 0.245 \\
200 & 0.226 & 0.182 & \textbf{0.339} & 0.264 \\
300 & 0.203 & 0.270 & \textbf{0.358} & 0.238 \\
400 & 0.268 & 0.262 & \textbf{0.319} & 0.228 \\
500 & \textbf{0.295} & 0.250 & 0.278 & 0.234 \\
600 & \textbf{0.292} & 0.286 & 0.283 & 0.250 \\
700 & 0.292 & \textbf{0.292} & 0.315 & 0.264 \\
800 & \SetCell[c=4]{c}{0.260} \\
\end{tblr}
\caption{Correlation of distances derived from feature subsets with $G_d$ for each selection method and subset size. Higher values indicate better performance. Best results for each subset size are highlighted.}
\label{tab:align}
\end{table}

Table \ref{tab:align} shows detailed results of distance alignment with modified Gower coefficient $G_d$ scores computed by \cite{hammarstrom-oconnor-2013-dependency}, for all four feature selection methods: Variance (Var), PCA Loadings (PCA), Laplacian Score (LS) and Correlation-based Feature Selection (CFS). Each feature subset, at its best, yields stronger alignment compared to the baseline. The 24 low-resource languages studied were: Sambu, Cayapa, Paya, Bintucua, Cagaba, Ulua, Paez, Muisca, Huaunana, Boruca, Misquito, Quiche, Lenca, Cuna, Xinca, Camsa, Cofan, Colorado, Cabecar, Bribri, Catio, Bocota, Teribe, and Movere.

\section{Downstream Tasks \label{appendix:downstream}}
\begin{figure*}[h!]
\centering
\includegraphics[width=1\textwidth]{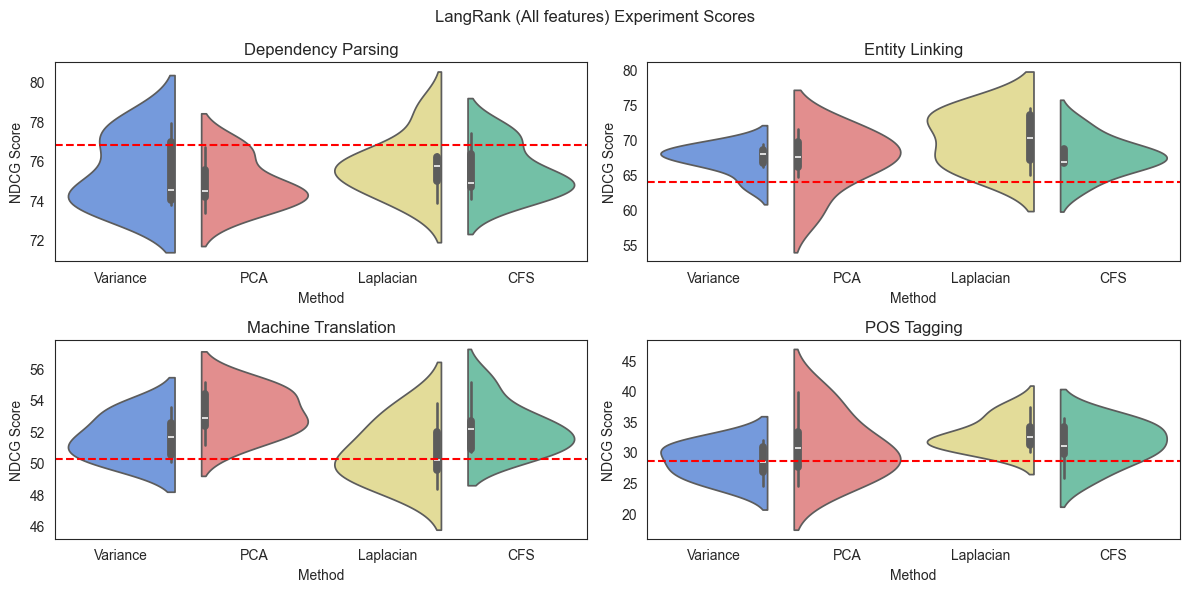}
\caption{LangRank (using all features) NDCG@3 scores across each sub-task. Dotted line indicates baseline performance when training with URIEL+ distances. \textbf{Higher is better.}}
\label{fig:langrank-all-violin}
\end{figure*}

\begin{figure*}[h!]
\centering
\includegraphics[width=1\textwidth]{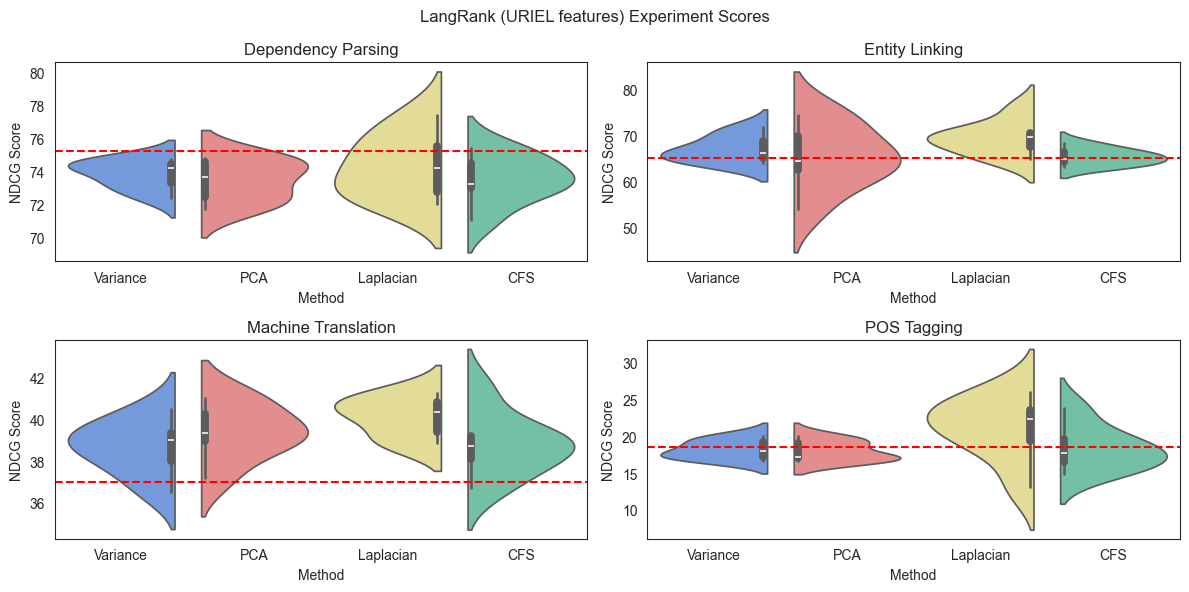}
\caption{LangRank (using only URIEL features) NDCG@3 scores across each sub-task. Dotted line indicates baseline performance when training with URIEL+ distances. \textbf{Higher is better.}}
\label{fig:langrank-uriel-violin}
\end{figure*}

\begin{figure*}[h!]
\centering
\includegraphics[width=1\textwidth]{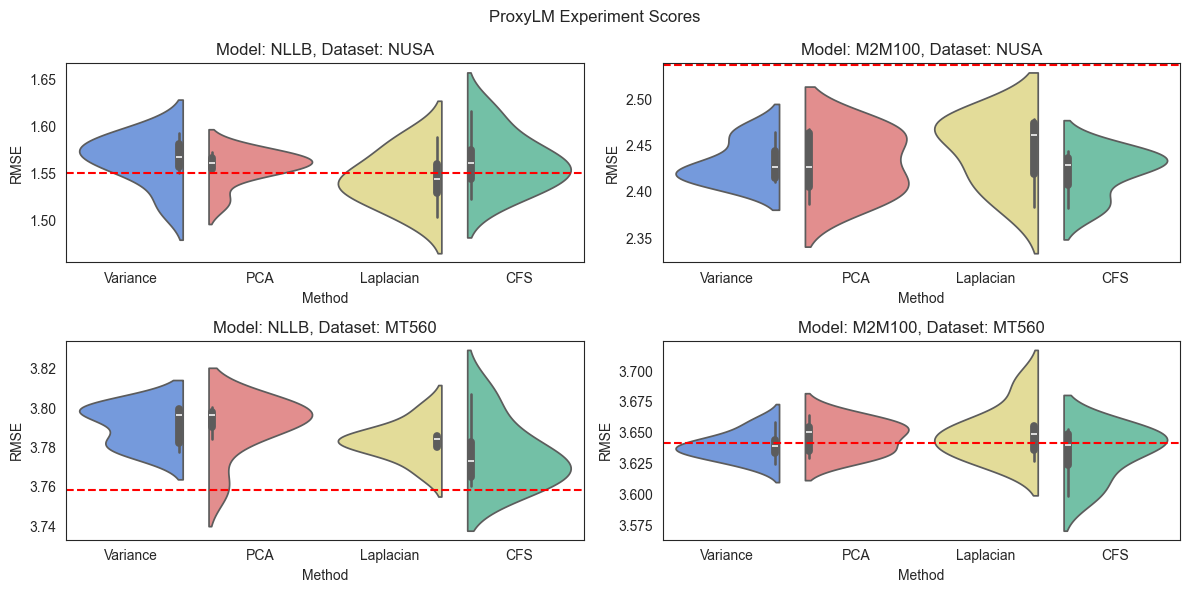}
\caption{ProxyLM RMSE scores across each model and sub-task. Dotted line indicates baseline performance when training with URIEL+ distances. \textbf{Lower is better.}}
\label{fig:proxy-violin}
\end{figure*}

\begin{figure*}[h!]
\centering
\includegraphics[width=1\textwidth]{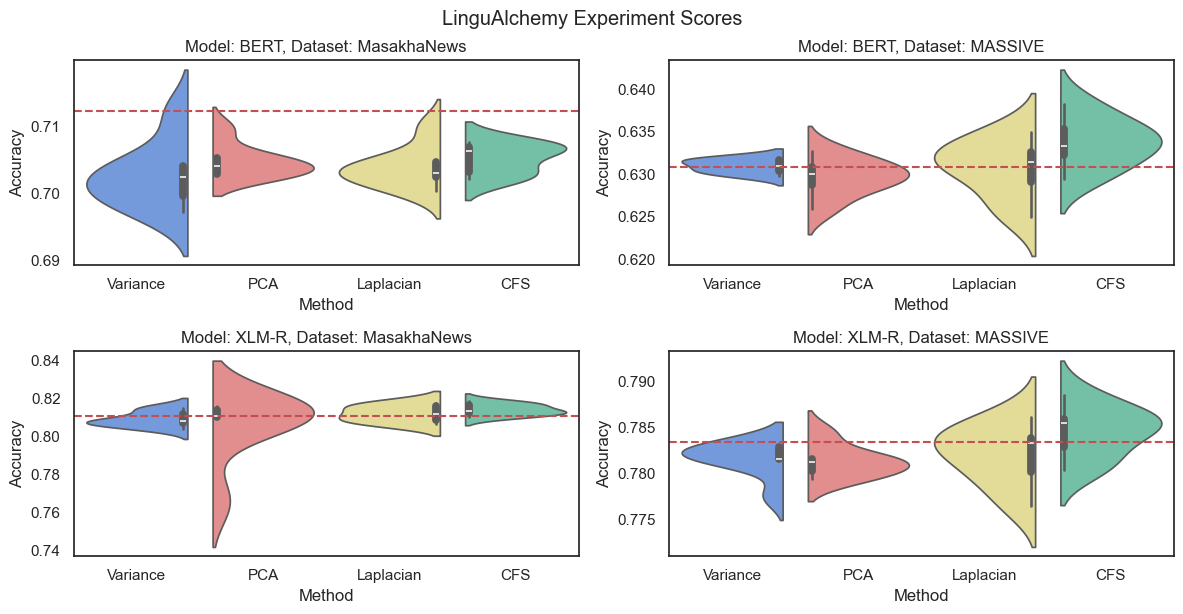}
\caption{LinguAlchemy accuracy scores across each model and sub-task. Dotted line indicates baseline performance with URIEL+'s syntax vectors. \textbf{Higher is better.}}
\label{fig:lingu-violin}
\end{figure*}
\subsection{Experimental Settings}
We ran \textsc{LangRank} with the same Glottocode replacements for some languages as detailed in \citet{khan-etal-2025-uriel-plus}. We evaluated the performance of the trained LightGBM ranker by its mean Normalized Discounted Cumulative Gain score with $k=3$, which measures the ranking quality of the three highest rankings, across cross-validation folds. Since LangRank uses linguistic features from a variety of datasets, we performed our experiment in two settings: (1) using only typological distance features and (2) using all features.

We used \textsc{ProxyLM} to train XGBoost regressors under the random test-split setting on model performance in both the English-centric and Many-to-many-language datasets, using the same regressor hyperparameters as in \citet{anugraha2024proxylm}. The performance of the regressor is determined by its root mean squared error (RMSE), which measures the difference between \textsc{ProxyLM}'s predicted and actual values.

We replicated the \textsc{LinguAlchemy} pipeline used by \citet{khan-etal-2025-uriel-plus} for evaluating URIEL+'s syntax vectors. For each feature subset, we extracted syntax vectors by taking only the syntax features (features whose names begin with "S\_") from the imputed, reduced-dimensionality vectors. For comparison with these syntax vectors, we applied URIEL+'s imputed syntax vectors as the baseline.

For evaluation, we computed average improvement by taking the average percentage change (or negative percentage change for \textsc{ProxyLM} experiments, as lower RMSE represents an improvement) in experiment scores, compared to the baseline, across all sub-tasks. No GPU was required to run either \textsc{LangRank} and \textsc{ProxyLM}, while all \textsc{LinguAlchemy} experiments were run on a single H100, utilizing 200 compute hours.

\subsection{Detailed Results}
Figures \ref{fig:langrank-all-violin}, \ref{fig:langrank-uriel-violin}, \ref{fig:proxy-violin} and \ref{fig:lingu-violin} show detailed experiment scores across sub-tasks for all feature subsets. For each feature selection method, evaluation scores across subset sizes are aggregated in a single distribution plot.

\section{Reproducibility and Licenses}
\label{appendix:reproduce}
All experiments were conducted with publicly available data. To support transparency and reproducibility, our code can be found at: \url{https://github.com/Swithord/URIELPlusOptimization}. In particular, our method uses the PCA implementation in \texttt{scikit-learn (v1.5)} \cite{scikit-learn}, along with the SoftImpute implementation in \texttt{fancyimpute (v0.7)} \cite{fancyimpute} with the mean initial fill method.

The URIEL+ database, along with evaluation frameworks \textsc{ProxyLM}, \textsc{LinguAlchemy} and \textsc{LangRank}, are licensed under CC BY 4.0.
\end{document}